\documentclass[10pt,twocolumn,letterpaper]{article}

\usepackage{cvpr}              
\usepackage[accsupp]{axessibility}
\usepackage{multirow}
\usepackage[utf8]{inputenc} 
\usepackage{xcolor}
\usepackage{float}
\definecolor{cvprblue}{rgb}{0.21,0.49,0.74}
\usepackage[pagebackref,breaklinks,colorlinks,allcolors=cvprblue]{hyperref}
\def\paperID{14} 
\def\confName{EarthVision}
\def\confYear{2026}

\title{SatUnreal: A High-Precision Synthetic Dataset \\for Satellite Stereo Matching via Unreal Engine}

\author{
Han-Gyeol Kim$^*$ \quad 
JaeWan Park$^*$ \quad 
Junmin Park \quad 
Darongsae Kwon \\
TelePIX, South Korea \\
{\tt\small \{khanai, eric.park, jmp\_TA, darong.kwon\}@telepix.net}
}

\begin{document}
\maketitle
\let\thefootnote\relax\footnotetext{$^*$Equal contribution.}
\begin{abstract}
3D reconstruction from satellite imagery is essential for large-scale topographic analysis, yet the lack of high-fidelity training datasets with accurate occlusion labels remains a primary bottleneck. Existing benchmarks, such as US3D and WHU-Stereo, face inherent challenges in spatio-temporal mismatch---environmental changes and shadow displacements between multi-view acquisitions---and provide ambiguous ground truth in occluded regions due to LiDAR sparsity. 

In this paper, we propose SatUnreal, a high-precision synthetic dataset designed to fundamentally overcome these limitations through an Unreal Engine-based simulation pipeline. SatUnreal provides 10,000 stereo pairs with high resolution ($0.3\text{m}$ GSD) and is characterized by: (1) Physical Geometry Simulation, replicating realistic satellite orbits by systematically varying baselines and azimuths; (2) Spatio-temporal Consistency, eliminating temporal noise through fixed virtual environments; (3) Topographic Diversity, spanning dense urban canyons to low-texture natural terrains; and (4) Mathematical Label Integrity, utilizing a novel two-step linetrace algorithm to generate flawless occlusion masks.

Experimental results using SOTA iterative models demonstrate that models trained exclusively on SatUnreal achieve superior zero-shot transfer performance on real-world benchmarks (US3D, WHU-Stereo) compared to those trained on real datasets. Our findings prove that physically accurate synthetic data provides a more effective supervisory signal for learning geometric features than complex real-world observations, establishing a new paradigm for Sim-to-Real transfer in Earth Observation.
\end{abstract}    
\section{Introduction}
\label{sec:intro}

\begin{figure}[h]
\begin{center}
   \includegraphics[width=1.0\linewidth]{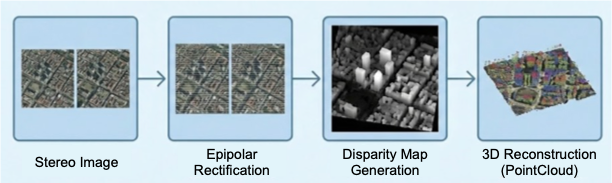} 
\end{center}
\vspace{-5mm}
   \caption{Process of 3D reconstruction from satellite images.}
\label{fig:proceeof3d} 
\end{figure}

With the rapid advancement of earth observation technologies and the increasing availability of high-resolution satellite imagery, 3D terrain reconstruction and urban modeling have become pivotal in various domains, including disaster management, urban planning, and high-definition mapping for autonomous navigation. In this context, generating disparity maps through stereo matching is the fundamental and most critical step in the 3D reconstruction pipeline. \Cref{fig:proceeof3d} shows the pipeline. While recent deep learning-based models have demonstrated performance surpassing human capabilities on terrestrial datasets such as Middlebury \cite{scharstein2014high} and KITTI \cite{menze2015joint}, their application to the satellite imagery domain remains a significant challenge.

The fundamental bottleneck in satellite stereo matching lies in the absence of high-quality training datasets. While existing benchmarks like US3D and WHU-Stereo represent extensive efforts, generating perfectly aligned ground truth from complex real-world acquisitions remains challenging. First is the spatio-temporal mismatch. Although modern agile satellites can mitigate this via in-orbit (along-track) stereo, acquiring such single-pass imagery at scale remains prohibitively expensive. Consequently, large datasets frequently rely on multi-date acquisitions, where inevitable time lags introduce supervisory noise like shifting shadows and terrain changes. Second is the inaccuracy of Ground Truth (GT). Even if stereo pairs are temporally perfectly aligned, asynchronous terrain data (e.g., LiDAR) inherently misaligns with satellite imagery, causing severe errors at object boundaries and in occluded regions \cite{lindenbergh2022spatio}.

To address these issues, research on synthetic data utilizing virtual engines has gained significant momentum. Wood \etal \cite{wood2021fake} demonstrated that models can achieve outstanding performance in real-world environments using only synthetic data, and Onishi \etal \cite{onishi2022impact} confirmed its effectiveness in compensating for the lack of topographic diversity. Synthetic data offers the distinct advantage of providing physically perfectly aligned labels and allowing for the simulation of extreme viewing angles and diverse terrain conditions that are difficult to capture in the real world.

In this paper, we propose SatUnreal, a high-precision synthetic satellite stereo dataset created using Unreal Engine to fundamentally overcome the limitations of real-world data. Moving beyond simple image synthesis, SatUnreal precisely replicates the physical geometric structure of satellites. Our primary contributions are as follows:
\begin{itemize}
    \item \textbf{Geometric Similitude \& Multi-Azimuth Simulation:} Using a scaled aerial proxy to mathematically replicate satellite orbital dynamics, we introduce a baseline rotation mechanism to simulate diverse acquisition azimuths. This ensures comprehensive coverage of complex urban structures often occluded in single-heading datasets.
    \item \textbf{Mathematical Label Extraction:} We designed a novel two-step linetrace algorithm utilizing the engine's geometry buffer to extract mathematically flawless occlusion masks, directly addressing the primary source of 3D reconstruction errors in real-world data.
    \item \textbf{Topographic Diversity:} We ensured broad topographic adaptability by encompassing not only dense urban areas but also low-texture natural terrains such as deserts and plains.
    \item \textbf{Spatio-temporal Consistency:} Generating multi-view images and labels simultaneously within a static temporal frame fundamentally eliminates chronic mismatch issues, enabling SatUnreal-trained models to achieve superior zero-shot generalization on real-world imagery.
\end{itemize}
\section{Related Work}
\label{sec:related_work}

\subsection{Evolution from Traditional to Learning-based Stereo Matching}
Historically, satellite stereo matching relied on geometry-based optimization like Semi-Global Matching (SGM) \cite{Hirschmuller_2005, Banz_Pirsch_Blume_2012}. However, traditional methods struggle in low-texture regions and are highly sensitive to radiometric differences \cite{Yang_Ahuja_2012, Yang_Wang_Yang_Stewénius_Nistér_2009, Hirschmüller_Scharstein_2009, Facciolo_Franchis_Meinhardt_2017}. Consequently, the field rapidly shifted toward deep learning paradigms, achieving superior robustness by learning high-level features from massive data \cite{Zhou_Meng_Cheng_2020, Le_Ranzato_Monga_Devin_Corrado_Chen_Dean_Ng_2011}. As algorithms advance, the availability of high-quality training datasets has emerged as the primary bottleneck \cite{Kim_Choi_Ahn_Min_2024}.

\subsection{Satellite Stereo Datasets and Their Limitations}
The performance of satellite stereo matching is fundamentally governed by the quality and quantity of the training data. Early research primarily relied on datasets derived from real-world satellite imagery, such as US3D \cite{bosch2019semantic} and WHU-Stereo \cite{huang2022whu}. While US3D established a large-scale urban training environment by combining multi-view satellite imagery with LiDAR data, it remains a challenging task due to the spatio-temporal mismatch issues inherent in satellite revisit periods \cite{lindenbergh2022spatio}. Specifically, the temporal gap between stereo pairs leads to shadow displacement caused by shifting solar angles, creating situations where matching algorithms might inadvertently learn photometric noise rather than true geometric correspondences \cite{tatar2017effect}. 

\begin{figure}[h] 
  \centering
  \includegraphics[width=0.97\linewidth]{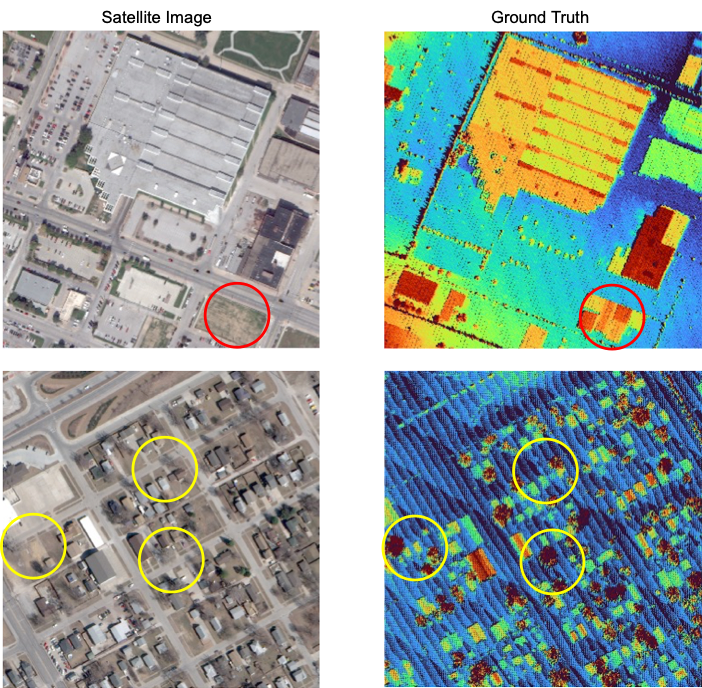}
  \caption{Temporal mismatch in existing datasets. The GT (right) often misaligns with the satellite image (left) due to time gaps.}
  \label{fig:temporal_mismatch}
\end{figure}

Furthermore, LiDAR-based GT frequently exhibits discrepancies ranging from months to years relative to the satellite acquisition time. Environmental changes during this interval—such as new constructions, demolitions, and vegetation growth—fundamentally undermine the reliability of the GT \cite{masquil2026diachronic}. A representative example of this discrepancy is illustrated in \cref{fig:temporal_mismatch}. Due to multi-year or seasonal temporal gaps between acquisitions, the LiDAR-derived GT often misrepresents the actual scene. For instance, an empty lot in the satellite image may appear as a fully constructed building in the GT (red circle), and bare winter trees can be incorrectly supervised as dense summer vegetation (yellow circles). Such significant mismatches can lead deep learning models to learn incorrect geometric correspondences, posing a challenge to preserving object boundaries \cite{che2022edge}.

\subsection{Advancement of Synthetic Data in Remote Sensing}
To circumvent the high costs and labeling errors of real-world datasets, synthetic data research utilizing virtual engines has emerged as a promising alternative. Over the past decade, there has been a consistent flow of synthetic datasets for generalized stereo matching (e.g., MPI Sintel \cite{butler2012naturalistic}, UnrealStereo4K \cite{8490973}, and recently Infinigen-based WMGStereo \cite{yan2025proceduraldatasetgenerationzeroshot}) and remote sensing (e.g., SynRS3D \cite{song2024synrs3d} and CARLA-based SkyScenes \cite{khose2023skyscenes}). Wood \etal \cite{wood2021fake} demonstrated that models trained exclusively on synthetic data could achieve high generalization performance in real-world domains, while Onishi \etal \cite{onishi2022impact} confirmed its effectiveness in compensating for topographic scarcity and maximizing data augmentation. 

While these existing synthetic pipelines are highly effective for their respective domains such as autonomous driving and drone-based aerial vision, adapting them directly for satellite photogrammetry presents a structural challenge. The primary bottleneck lies in finding a methodology capable of generating massive, high-density metropolitan datasets while rigorously adhering to the unique orbital geometry of Earth Observation satellites, which are characterized by near-orthographic projections and highly specific multi-azimuth convergence angles. Therefore, establishing a custom methodology that maximally preserves satellite geometry while enabling large-scale dataset generation remains a critical, unaddressed challenge in the field.

\begin{figure*}[!t]
    \centering
    \includegraphics[width=0.9\linewidth]{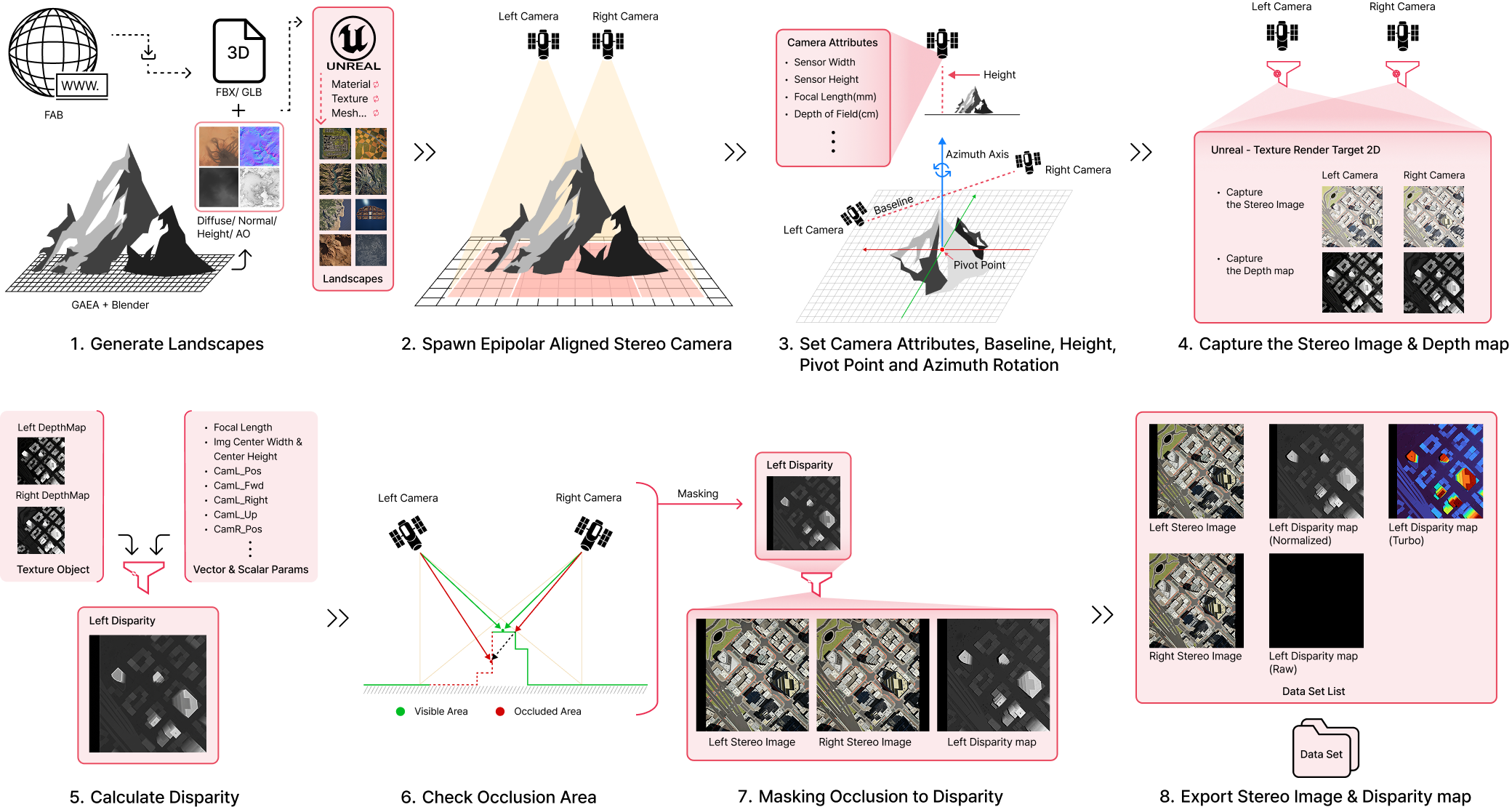}
    \caption{Overview of the SatUnreal dataset generation pipeline.}
    \label{fig:pipeline}
    \vspace{-2mm}
\end{figure*}

\subsection{Deep Learning for Stereo Matching}
Stereo matching has progressed significantly with the introduction of iterative update mechanisms such as RAFT-Stereo \cite{lipson2021raft} and IGEV \cite{xu2023iterative}. RAFT-Stereo refines disparity maps by repeatedly referencing global correlations, while IGEV effectively captures both low and high frequency details using a geometry encoding volume \cite{xu2023iterative}. 

In the satellite domain, attention-based models designed to extract global context, such as Selective-IGEV \cite{wang2024selective} and IAASNet \cite{huang2025iaasnet}, have become mainstream. Nevertheless, these sophisticated models still require clear supervision in fine structures. Inaccurate GT from real-world data hinders the convergence of these architectures. This study demonstrates how the mathematically flawless occlusion labels and high-resolution GT provided by SatUnreal unleash the full potential of SOTA deep learning architectures in the real-world domain.

Although recent foundation models exhibit strong zero-shot capabilities, they struggle with near-orthographic satellite geometry, making physically accurate synthetic datasets essential for domain-specific adaptation.
\section{SatUnreal Dataset Generation}
\label{sec:generation}

The overall pipeline for generating the SatUnreal dataset is illustrated in \cref{fig:pipeline}. The workflow is systematically designed from 3D asset configuration in Unreal Engine to mathematically precise label extraction, ensuring physically grounded and temporally consistent supervision.

\subsection{Virtual Environment and Assets Overview}
To ensure the robustness and generalization capability of the generated dataset, we utilized a comprehensive collection of 3D assets within the Unreal Engine environment, prioritizing high-fidelity urban models while incorporating diverse natural landscapes. 

\begin{figure}[h]
\centering
\includegraphics[width=\linewidth]{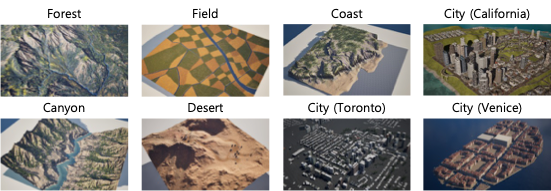}
\caption{Overview of 3D assets used for dataset generation.}
\label{fig:asset_overview}
\vspace{-3mm}
\end{figure}

As visualized in \cref{fig:asset_overview}, our urban models encompass various architectural styles and densities (e.g., California \cite{asset_california}, Toronto \cite{asset_toronto}, and Venice \cite{asset_venice}), acquired from the Fab Marketplace under academic research licenses. Complementing these, our custom-built natural environments---comprising procedurally generated canyons, deserts, forests, fields, and coastal areas---were specifically engineered to present distinct topographic and radiometric challenges.

Crucially, this asset selection was deliberately tailored as a purpose-driven stress-test for stereo matching networks. Rather than serving as mere aesthetic backgrounds, these environments provide controlled edge cases: natural landscapes like deserts and fields introduce severe low-texture ambiguities to challenge photometric consistency assumptions, while our high-density urban and canyon models induce extreme elevation changes and multiple occlusions. This setup rigorously evaluates the networks' capacity to preserve sharp depth discontinuities and handle complex geometric transitions that are often underrepresented in existing real-world datasets.

\subsection{Geometric Baseline and Scaled Simulation}

To replicate actual satellite imaging geometries within a localized simulation environment, we employ a systematic baseline scaling method based on a flat-earth approximation \cite{aguilar2022assessment}. Since our simulated urban models are confined to a localized region, we treat the orbital altitude ($H$) as a constant vertical distance from a planar ground surface.

While operational satellites typically orbit at altitudes of around 500\text{km}, generating such extreme ranges in a 3D engine to capture a limited area can lead to floating-point precision errors and a critical loss of overlapping regions necessary for stereo matching. Therefore, we deliberately scale our simulation geometry to an altitude $H$ of 1\text{km}. 

To ensure physical accuracy during rendering, we configured the virtual camera's focal length and sensor size (CCD) to match standard high-resolution aerial photogrammetry specifications. This aerial sensor profile acts as a mathematically rigorous proxy: by tuning these parameters to achieve a Ground Sample Distance (GSD) of 0.3\text{m} at a 1\text{km} altitude, the scaled model faithfully preserves the narrow Field-of-View (FOV) and orthographic-like projective geometry inherent in satellite sensors. The fundamental stereo geometry is maintained because it relies on the baseline-to-height ratio ($B/H$), rather than absolute distances.

In satellite stereo reconstruction, the convergence angle ($\theta$)---the angle between two optical axes---is a critical parameter that determines the trade-off between geometric triangulation precision and matching success rates \cite{aguilar2022assessment}. In the SatUnreal suite, all generated stereo pairs are subjected to epipolar rectification, aligning the optical axes to a nadir-viewing geometry. Under this configuration, the effective convergence angle ($\theta$) is defined as:
\begin{equation}
    \theta = 2 \cdot \arctan\left(\frac{B}{2H}\right)
\end{equation}

\begin{figure}[t]
  \centering
  \includegraphics[width=\linewidth]{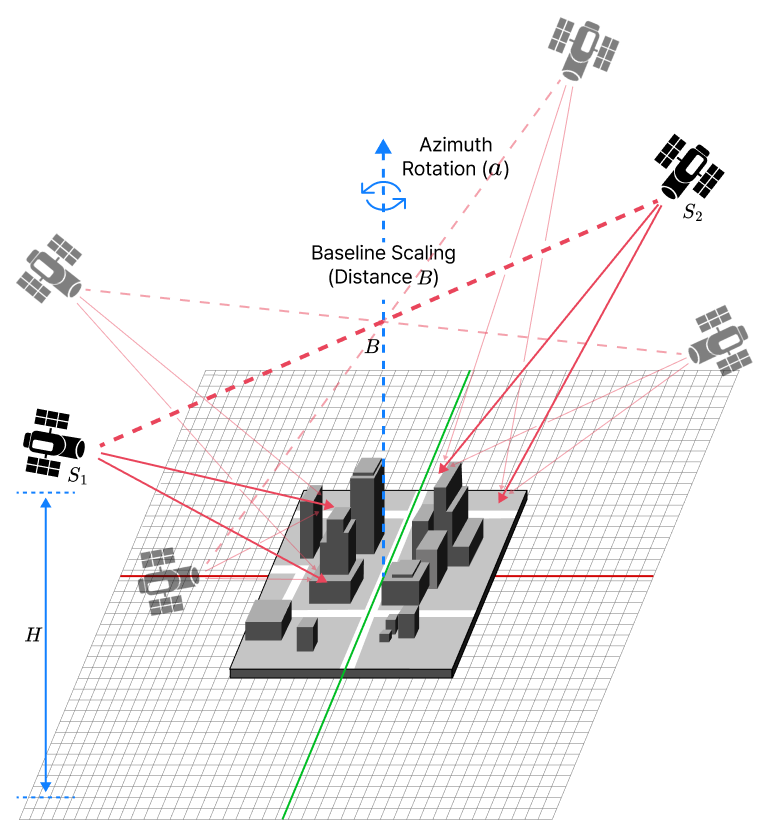}
  \caption{Schematic of the multi-azimuth stereo acquisition strategy. Under a scaled constant altitude ($H=1$km), geometric diversity is achieved via two mechanisms: (1) Baseline Scaling ($B$) to vary the effective convergence angle ($\theta$), and (2) Azimuth Rotation ($\alpha$) to capture structures from diverse orbital headings.}
  \vspace{-3mm}
  \label{fig:acquisition_strategy}
\end{figure}

To maximize geometric diversity, we introduce a controlled baseline rotation mechanism alongside scaling, as illustrated in \cref{fig:acquisition_strategy}. By varying the horizontal baseline $B$ from 50\text{m} to 200\text{m} at the fixed altitude $H = 1\text{km}$, we simulate convergence angles ranging from approximately 2.5$^\circ$ to 12.0$^\circ$. This specific range mirrors the optimal narrow-angle imaging conditions heavily favored for dense urban satellite acquisitions to minimize severe occlusions \cite{aguilar2022assessment}. Furthermore, we rotate the entire baseline apparatus around a central vertical axis to arbitrary azimuth angles ($\alpha$). This ensures robust coverage of complex urban structures, such as building facades, from diverse orbital headings.

The final dataset is produced at a very high resolution of 0.3\text{m}, matching the specifications of state-of-the-art commercial satellites such as WorldView-3. Each image is generated at 2048 $\times$ 2048 pixels, covering a ground area of approximately 614.4\text{m} $\times$ 614.4\text{m}. This high-resolution design provides exceptional versatility, enabling research applications ranging from ultra-precise urban 3D modeling to wide-area terrain analysis via downsampling.

While scaling the simulation altitude to 1km preserves the geometric Baseline-to-Height (B/H) ratio and circumvents the floating-point precision limits of 3D engines, it introduces a specific radiometric downside. Complex atmospheric effects—such as Rayleigh scattering and haze naturally occurring over a 500km column—must be vertically compressed into a 1km space. Despite utilizing engine approximations, this artificial condensation leaves a slight radiometric domain gap compared to true orbital imagery.

\subsection{Clean Spatio-temporal Supervision}

Complementing the precise geometric control described above, SatUnreal addresses fundamental limitations inherent in real-world satellite datasets caused by temporal acquisition gaps. Conventional real-world satellite datasets often suffer from inconsistent supervision signals due to transient phenomena like shifting shadows, moving objects, or structural changes between asynchronous stereo captures.

SatUnreal resolves these issues through a strictly controlled simulation environment that guarantees perfect spatio-temporal consistency. By freezing the entire state—including illumination conditions, atmospheric effects, and dynamic object positions—during the multi-view generation process, we ensure that the only significant variation between stereo pairs is the camera viewing angle. This approach isolates geometric disparity as the sole variable, effectively eliminating environmental noise and providing highly trustworthy GT. Consequently, this clean supervision signal enables learning-based models to focus exclusively on geometric matching, leading to improved convergence speed and overall reconstruction precision \cite{huang2025iaasnet}.

\subsection{High-Precision Occlusion and Disparity}
Occlusion caused by vertical walls or abrupt terrain changes is a chronic issue in satellite stereo matching. While existing datasets rely on estimation \cite{bosch2019semantic}, SatUnreal generates mathematically flawless labels by leveraging the internal View-Projection (VP) matrices of the rendering engine.

\subsubsection{Matrix-based Disparity via Reprojection}
To achieve sub-pixel precision and guarantee rigorous geometric consistency across varied baseline configurations, we utilize a reprojection-based approach instead of simplified geometric approximations:

First, a 3D point $P_w$ in the world space is back-projected from the left image pixel $(u_L, v_L)$ using its corresponding depth $Z$ and the inverse of the left View-Projection matrix ($VP_L^{-1}$):
\begin{equation}
    P_w = VP_L^{-1} \cdot \mathbf{x}_{NDC}(u_L, v_L, Z)
\end{equation}
where $\mathbf{x}_{NDC}$ denotes the pixel coordinates in the Normalized Device Coordinate space.

Next, $P_w$ is projected onto the right image plane using the right camera's VP matrix ($VP_R$) to determine its theoretical coordinate $(u_R, v_R)$:
\begin{equation}
    \mathbf{x}_R(u_R, v_R) = VP_R \cdot P_w
\end{equation}

Finally, the disparity $\mathbf{d}$ is extracted as the displacement vector between the two coordinates:
\begin{equation}
    \mathbf{d} = (u_L, v_L) - (u_R, v_R)
\end{equation}
Since all stereo pairs in SatUnreal are strictly epipolar-rectified, the vertical displacement is inherently zero ($v_L = v_R$). Consequently, the displacement vector $\mathbf{d}$ naturally simplifies to the 1D horizontal disparity $d = u_L - u_R$.
This approach ensures that the generated disparity labels are mathematically derived directly from the engine's precise camera parameters, devoid of any estimation errors.

\subsubsection{Two-Step LineTrace for Occlusion Masking}
\begin{figure}[h]
\begin{center}
   \includegraphics[width=\linewidth]{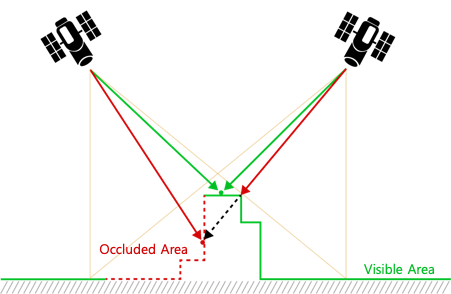} 
\end{center}
\vspace{-5mm}
   \caption{Occlusion in satellite stereo imaging. The solid green line represents the terrain visible to the right sensor. The dashed red line indicates the occluded region(visible to the left sensor but blocked from the right).}
   \vspace{-3mm}
\label{fig:Occlusion} 
\end{figure}

To complement the disparity generation, we propose a two-step linetrace algorithm, as illustrated in \cref{fig:Occlusion}, to identify valid correspondences:
\begin{itemize}
    \item \textbf{Step 1 (Visibility Check):} Rays are cast from the left camera to identify collision points ($P_w$) with the surface and objects.
    \item \textbf{Step 2 (Path Verification):} A reverse ray is cast from $P_w$ to the right camera ($C_R$). If a collision occurs before reaching $C_R$, the pixel is determined to be occluded and marked as invalid.
\end{itemize}

Rather than relying on conventional Z-buffer depth thresholding, which often fails in multi-layered architectural environments, our two-step linetrace algorithm operates directly in the 3D world space using continuous bidirectional ray-casting. This mathematically guarantees physical visibility, effectively distinguishing between self-occlusions in complex urban canyons and valid multi-layered geometries. Consequently, it yields strictly binary, flawless occlusion masks that prevent deep learning models from hallucinating correspondences in unobservable regions.

\subsubsection{Dataset Statistics and Composition}

The SatUnreal dataset comprises a comprehensive collection of 10,000 high-resolution stereo pairs. As detailed in \cref{tab:dataset_stats}, this scale is achieved through a systematic combinatorial acquisition strategy: for each of the selected pivot points across diverse virtual environments, we capture 20 linear baseline steps (ranging from 50\text{m} to 200\text{m}) and 20 random azimuth rotation angles ($\alpha \in [-90^\circ, 90^\circ]$). This rigorous parameter sampling deliberately concentrates the effective convergence angles within the optimal narrow-angle spectrum ($2.5^\circ$ to $12.0^\circ$) while ensuring robust multi-directional coverage of 3D structures. 

\begin{figure}[t]
\centering
\includegraphics[width=1.0\linewidth]{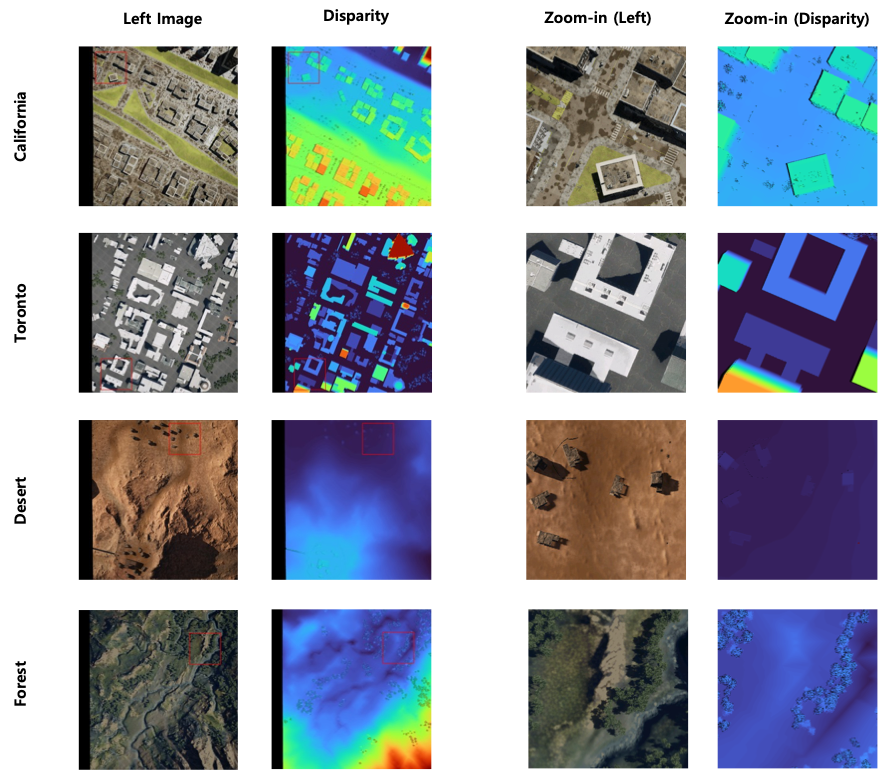}
\caption{Qualitative visual samples of SatUnreal across various domains. The columns present the image, Disparity GT, and their respective zoom-in patches. The zoom-in patches highlight the mathematically precise alignment and boundary preservation.}
\label{fig:visual_samples}
\vspace{-3mm}
\end{figure}

Scenario-wise, the dataset maintains a well-curated balance. Approximately 80.0\% (8,000 pairs) is dedicated to complex urban environments (California, Toronto and Venice models) to challenge models with severe occlusions and abrupt depth discontinuities. The remaining 20.0\% (2,000 pairs) covers diverse natural landscapes (canyons, deserts, forests, fields, and coasts) to rigorously evaluate model generalizability across varying texture densities and elevation changes. Qualitative visual samples of the generated dataset across these diverse domains are presented in \cref{fig:visual_samples}.

Furthermore, to emulate the varying degrees of stereo overlap inherent in agile satellite acquisitions, SatUnreal preserves full-frame imagery while explicitly encoding non-overlapping regions with null values. This design offers dual-mode training flexibility: models can either be trained directly on uncropped frames to develop robustness against low-overlap conditions, or dynamically cropped using the null-value masks to extract only the common overlapping regions. This latter approach ensures seamless structural compatibility and data fusion with legacy benchmarks that predominantly utilize pre-cropped patches.

\begin{table}[h]
\centering
\caption{Quantitative summary of the SatUnreal dataset. Note: All configurations use 20 baseline steps (50--200 m) and 20 random azimuths $\alpha \in [-90^\circ, 90^\circ]$.}
\label{tab:dataset_stats}
\resizebox{\linewidth}{!}{%
\begin{tabular}{lcc @{\hspace{3em}} lcc}
\toprule
\multicolumn{3}{c}{\textbf{Urban Environments}} & \multicolumn{3}{c}{\textbf{Natural Landscapes}} \\
\cmidrule(r){1-3} \cmidrule(l){4-6}
\textbf{Region} & \textbf{Pivot Points} & \textbf{Total Pairs} & \textbf{Region} & \textbf{Pivot Points} & \textbf{Total Pairs} \\
\midrule
California & 8 & 3,200 & Forest & 1 & 400 \\
Toronto & 10 & 4,000 & Canyon & 1 & 400 \\
Venice & 2 & 800 & Coast & 1 & 400 \\
 & & & Desert & 1 & 400 \\
 & & & Field & 1 & 400 \\
\midrule
\multicolumn{6}{c}{\textbf{Overall Total: 25 Pivot Points, 10,000 Pairs}} \\
\bottomrule
\end{tabular}%
}
\end{table}

\section{Experiments}
\label{sec:experiments}

In this section, we evaluate the effectiveness of the proposed SatUnreal dataset through comprehensive quantitative and qualitative analyses. Our primary focus is investigating whether a model trained solely on high-fidelity synthetic data can achieve superior zero-shot transfer performance on real-world satellite imagery compared to models trained on existing real-world datasets \cite{wood2021fake, wang2024selective}.

\subsection{Experimental Setup}
\label{subsec:setup}
\begin{itemize}
\item \textbf{Models:} We utilize Selective-IGEV \cite{wang2024selective}, RAFT-Stereo \cite{lipson2021raft}, and DLNR \cite{10203487}, representing leading paradigms of recurrent iterative optimization and deep cascaded refinement well-suited for fine-grained terrain recovery.

\item{
\textbf{Baselines:} For a comparative study, we employ the most widely used real-world satellite datasets: US3D \cite{bosch2019semantic} and WHU-Stereo \cite{huang2022whu}. To ensure a fair comparison regarding dataset scale, we explicitly detail their splits: US3D contains 4,292 pairs (2,000 train, 292 val, and 2,000 test), while WHU-Stereo provides 10,934 patches (7,546 train, 770 val, and 2,618 test). In strict alignment, our SatUnreal dataset encompasses 10,000 pairs partitioned into 6,646 train, 500 val, and 2,854 test samples. This comparable training volume confirms that our zero-shot performance gains stem from high-fidelity geometric supervision rather than merely scaling the data size.}

\item \textbf{Training Protocol:} All models are initialized with weights pre-trained on SceneFlow \cite{mayer2016large}---a standard large-scale synthetic dataset for generic stereo matching---and fine-tuned for 50k iterations. Crucially, as our linetrace algorithm explicitly encodes unobservable regions with $0$, we apply a strict validity mask ($d > 0$) during loss computation.

\item \textbf{Evaluation Normalization for Zero-shot Transfer:} To ensure fair zero-shot comparisons, we align with the pre-cropped protocols of real-world datasets by dynamically cropping non-overlapping zero-padding regions in the SatUnreal set and normalizing disparity scales. This guarantees that quantitative metrics reflect true geometric matching capabilities rather than penalizing unconstrained spatial boundaries.

\item{
\textbf{Evaluation Metrics:} To rigorously assess the geometric accuracy, we employ three standard metrics: (1) End-Point Error (EPE), representing the average Euclidean distance between the predicted and ground-truth disparity; and (2) n-pixel Error Rates (1px and 3px), which denote the percentage of pixels whose absolute error falls within a threshold of 1 and 3 pixels, respectively.}

\end{itemize}
\subsection{Quantitative Results}
\label{subsec:quant}
The quantitative performance is summarized in \cref{tab:comprehensive_results}. Models trained purely on our synthetic SatUnreal dataset demonstrate highly competitive zero-shot generalization across established real-world benchmarks (US3D and WHU-Stereo). While models trained strictly in-domain naturally exhibit the lowest localized error rates on their respective test sets, the SatUnreal-trained models successfully bridge the Sim-to-Real gap, achieving comparable sub-pixel accuracy without exposure to real-world domain shifts during training

Importantly, SatUnreal provides a mathematically precise supervisory signal that prevents networks from overfitting to domain-specific artifacts, allowing them to maintain robust error bounds across diverse domains. As demonstrated in cross-dataset evaluations, models trained on SatUnreal show remarkable structural stability, successfully transferring learned geometric features to the complex distributions of real-world satellite imagery.

Furthermore, we observe that certain high-capacity architectures, such as Selective-IGEV, exhibit heightened sensitivity to spatial misalignments between image features and LiDAR-derived ground truths. When the mathematically optimized structural boundaries of the model encounter inherent Sim-to-Real mismatches, it can trigger disproportionately large localized errors. This severely inflates the global EPE, even though the models maintain high qualitative structural fidelity and preserve sharp depth discontinuities. This not only reaffirms that our synthetic data generation pipeline provides a structurally effective supervisory signal for Sim-to-Real transfer, but also highlights the architecture-specific dynamics when evaluating on complex real-world observations.

\begin{table*}[t]
\caption{Comprehensive quantitative results on synthetic and real-world benchmarks. We evaluate the internal consistency of SatUnreal and its zero-shot transferability to US3D and WHU-Stereo.}
\label{tab:comprehensive_results}
\vspace{-2mm}
\centering
\footnotesize 
\renewcommand{\arraystretch}{1.0}
\resizebox{\linewidth}{!}{
\begin{tabular}{l|c|ccc|ccc|ccc}
\hline
\multirow{3}{*}{Model} & \multirow{3}{*}{Train Set} & \multicolumn{9}{c}{Test Set} \\ \cline{3-11} 
 &  & \multicolumn{3}{c|}{SatUnreal} & \multicolumn{3}{c|}{US3D} & \multicolumn{3}{c}{WHU-Stereo} \\ \cline{3-11} 
 &  & EPE $\downarrow$ & 1px $\uparrow$ & 3px $\uparrow$ & EPE $\downarrow$ & 1px $\uparrow$ & 3px $\uparrow$ & EPE $\downarrow$ & 1px $\uparrow$ & 3px $\uparrow$ \\ \hline\hline
\multirow{3}{*}{RAFT-Stereo} 
 & US3D & 294.19 & 7.06 & 12.36 & 11.76 & 38.60 & 67.70 & 6.08 & 87.23 & 92.59 \\
 & WHU & 21.43 & 60.41 & 75.34 & 3.81 & \textbf{55.20} & \textbf{89.00} & \textbf{0.19} & \textbf{97.66} & \textbf{99.29} \\
 & SatUnreal & \textbf{0.57} & \textbf{90.78} & \textbf{98.16} & \textbf{2.45} & 48.83 & 86.87 & 1.40 & 87.48 & 96.36 \\ \hline
\multirow{3}{*}{Selective-IGEV} 
 & US3D & 6.30 & 15.11 & 75.28 & \textbf{1.00} & \textbf{64.30} & \textbf{95.12} & 2.18 & 24.19 & 94.12 \\
 & WHU & \textbf{2.64} & \textbf{84.52} & \textbf{95.09} & 46.13 & 14.46 & 29.32 & \textbf{0.18} & \textbf{98.25} & \textbf{99.35} \\
 & SatUnreal & 3.12 & 84.41 & 92.13 & 60.89 & 19.33 & 53.38 & 0.93 & 88.44 & 97.52 \\ \hline
\multirow{3}{*}{DLNR} 
 & US3D & 2.92 & 54.32 & 87.26 & \textbf{3.49} & \textbf{28.12} & \textbf{62.51} & 2.19 & 73.76 & 95.41 \\
 & WHU & 35.27 & 10.15 & 12.34 & 35.19 & 7.50 & 15.49 & \textbf{0.18} & \textbf{98.36} & \textbf{99.42} \\
 & SatUnreal & \textbf{0.26} & \textbf{93.74} & \textbf{99.36} & 3.84 & 22.03 & 60.79 & 1.33 & 82.09 & 96.78 \\ \hline
\end{tabular}
}
\end{table*}

\begin{figure*}[t]
\centering
\includegraphics[width=1.0\linewidth]{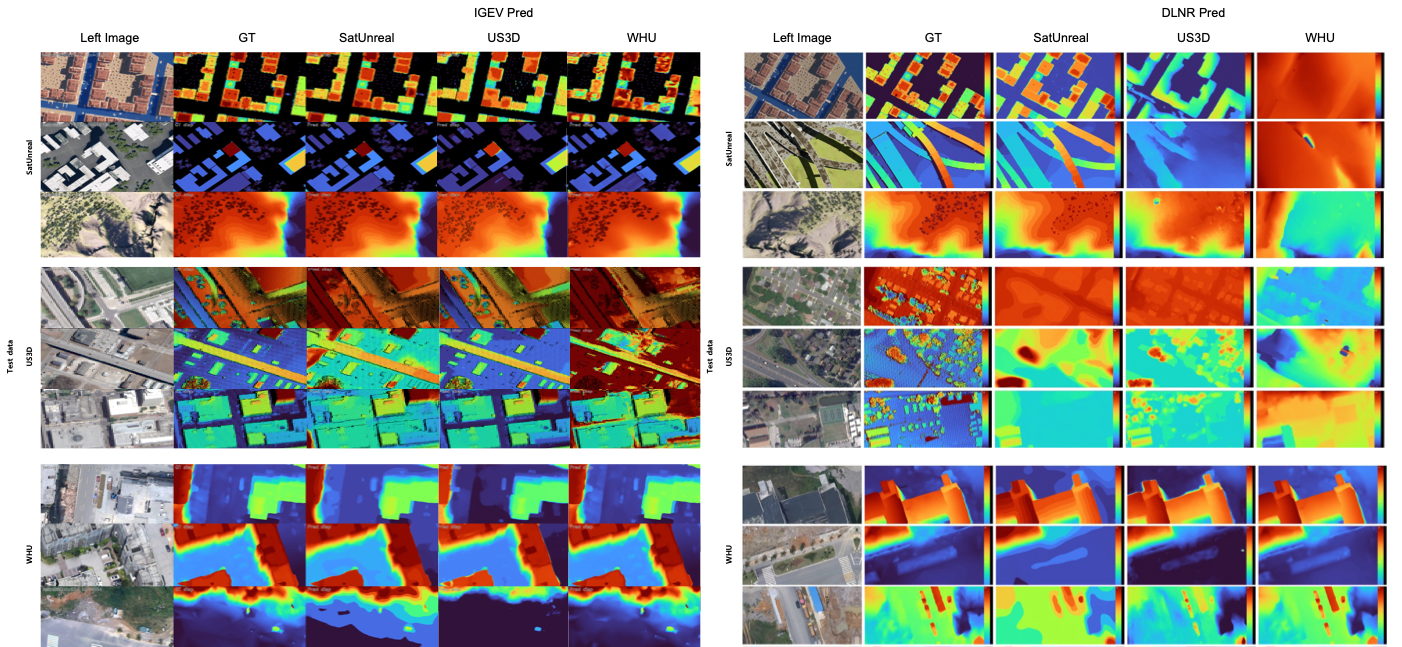}
\caption{Comprehensive cross-dataset qualitative evaluation using Selective-IGEV (Left) and DLNR (Right). The rows represent the test datasets (Top: SatUnreal, Middle: US3D, Bottom: WHU-Stereo). In each block, columns represent the Left image, GT, and predictions from models trained exclusively on SatUnreal, US3D, and WHU-Stereo. Left (Selective-IGEV): The model exhibits robust cross-domain compatibility. Notably, the SatUnreal-trained model consistently maintains clear architectural boundaries and effectively handles complex occlusions across all test domains. Right (DLNR): This highlights specific architecture-data dynamics; DLNR's global smoothing mechanism contrasts with the high-frequency variations often found in US3D's ground truths, resulting in predictions that prioritize global smoothness. Overall, our SatUnreal-trained models successfully preserve broad topographic structures across unseen domains.}
\label{fig:visual_comp}
\vspace{-3mm}
\end{figure*}

\subsection{Qualitative Analysis}
\begin{itemize}
\item \textbf{Occlusion Handling:} The efficacy of our explicit masking strategy is visually evident in complex urban canyons. While models trained on baseline datasets can face challenges with distinct boundary definitions around vertical walls, SatUnreal-trained models successfully maintain crisp, mathematically precise disparity discontinuities. This confirms that our two-step linetrace algorithm provides a robust supervisory signal for learning sharp geometric transitions.
\item \textbf{Cross-Domain Compatibility:} Qualitatively, we observe a remarkable degree of cross-domain compatibility among all models across the three datasets. Models trained on US3D and WHU-Stereo exhibit reasonable generalization to unseen domains, successfully retaining general geometric structures despite significant domain shifts. Notably, despite being trained purely on synthetic data, our models yield reconstruction qualities that are visually on-par with, and structurally comparable to, those trained on large-scale real-world datasets. This visually substantiates that the precise supervision of SatUnreal effectively bridges the Sim-to-Real gap without relying solely on extensive real-world annotations. Detailed visual comparisons across all domains are provided in the Supplementary Material.
\end{itemize}

\subsection{Discussion and Limitations}
We focus on zero-shot evaluation to strictly isolate SatUnreal's pure geometric supervisory quality, leaving real-world fine-tuning as a promising future step. Furthermore, our dataset size (10,000 pairs) ensures an equitable computational comparison with baselines, avoiding brute-force scaling, though our pipeline supports near-infinite expansion. Additionally, cross-domain training occasionally outperforms in-domain training (\cref{tab:comprehensive_results}). Rather than an anomaly, this highlights intricate optimization dynamics: noisy, complex datasets like US3D hinder highly iterative models from converging on generalized priors. Conversely, tightly curated data (WHU) or mathematically flawless synthetic data (SatUnreal) provide stable geometric features. Furthermore, real-world datasets act as essential texture regularizers for high-capacity architectures (e.g., Selective-IGEV), preventing synthetic overfitting and occasionally improving generalization even on synthetic test sets.
\section{Conclusion and Future Work}
\label{sec:conclusion}
We proposed SatUnreal, a high-precision synthetic dataset utilizing Unreal Engine to resolve spatio-temporal mismatches and occlusion ambiguities in satellite stereo benchmarks \cite{bosch2019semantic}. Through scaled geometric similitude, we achieved high practical fidelity. Although actual push-broom sensors exhibit non-linear epipolar curves, orbital stability renders these negligible for localized patches, validating our piecewise-linear approximations. Furthermore, our simulation guarantees multi-view consistency and provides mathematically precise occlusion masks via a two-step linetrace algorithm \cite{aguilar2022assessment}.

Experiments confirm that SatUnreal-trained models achieve competitive zero-shot transfer on real-world imagery, proving that physically precise synthetic data effectively mitigates real-world supervisory noise.

To further close the Sim-to-Real gap, future work will expand the dataset with dynamic illuminations, diverse biomes, and atmospheric scattering. Geometrically, we will integrate rigorous push-broom simulations (dynamic Rational Polynomial Coefficients, non-linear trajectories) into the rendering pipeline and support arbitrary elevation datums ($d<0$). The dataset, generation pipeline, and pre-trained models will be publicly released to advance robust satellite stereo matching.
{
    \small
    \bibliographystyle{ieeenat_fullname}
    \bibliography{main}
}


\end{document}